\documentclass[10pt,final,journal]{IEEEtran}
\usepackage[ruled,vlined,linesnumbered,resetcount]{algorithm2e}
\usepackage[nocompress]{cite}
\usepackage{graphicx,epstopdf,bm,amsthm,amsmath,marvosym,nopageno,mathtools,amssymb,color,url,environ,array,tabularx,svg}
\usepackage{environ,enumerate,multirow}
\renewcommand{\theenumi}{\alph{enumi}}

\graphicspath{{./figs/}}

\renewcommand{\v}[1]{\boldsymbol{\mathbf{#1}}} % vector notation
\newcommand{\opn}[1]{\operatorname{#1}} % operator name
\newcommand{\set}[1]{\mathbb{#1}} % blackboard boldface character (set notation)
\newcommand{\prt}[1]{\left(#1\right)} % parentheses
\newcommand{\sqb}[1]{\left[#1\right]} % square brackets
\newcommand{\expt}[1]{\mathbb{E}\left\{#1\right\}} % expectation operator
\newcommand{\inc}[1]{\in \set{C}^{#1}} % belong to the complex set of dimension #1
\NewEnviron{meq}{
    \begin{align}
        \begin{split}
            \BODY
        \end{split}
    \end{align}
    }

\begin{document}

\title{Direction Finding with Sparse Arrays Based on Variable Window Size Spatial Smoothing}

\author{Wesley S. Leite, Rodrigo C. de Lamare, Yuriy Zakharov, Wei Liu and Martin Haardt \vspace{-0.05em}

}

\maketitle

\begin{abstract}
In this work, we introduce a \emph{variable window size} (VWS) spatial smoothing framework that enhances coarray‑based direction of arrival (DOA) estimation for sparse linear arrays. By compressing the smoothing aperture, the proposed VWS Coarray MUSIC (VWS‑CA‑MUSIC) and VWS Coarray root-MUSIC (VWS‑CA‑rMUSIC) algorithms replace part of the perturbed rank‑one outer products in the smoothed coarray data with \emph{unperturbed} low‑rank additional terms, increasing the separation between signal and noise subspaces, while preserving the signal subspace span. We also derive the bounds that guarantees identifiability, by limiting the values that can be assumed by the compression parameter. Simulations with sparse geometries reveal significant performance improvements and complexity savings relative to the fixed‑window coarray MUSIC method. 
\end{abstract}


\IEEEpeerreviewmaketitle

\section{Introduction}

\IEEEPARstart{D}{irection‑of‑arrival} (DOA) estimation is of fundamental importance in modern communications, radar, sonar and radio‑astronomy. From the literature, it is known that its accuracy degrades severely when the impinging signals sources are mutually correlated or coherent\cite{tuncer2009,hassen2011,yang2019}. Even for uncorrelated sources, a designer has to deal with virtually coherent sources, once covariance‑vectorization strategies are applied to increase the degrees of freedom (DOF) of sparse linear arrays (SLAs), i.e., to estimate more sources than the number of physical sensors \cite{VanTrees2002,Pal2010-1,Leite2021,imisc,ssub,emisc}.

Early techniques focused on covariance matrix preprocessing such as forward spatial smoothing (FSS) and its forward–backward variant (FBSS), thus enabling subspace methods that include low-rank techniques \cite{jidf,jio,sjidf,jiostap,dce,jiodoa,alrdoa,stapadmm,lrcc,rdstap}, compressive sensing,  MUSIC, ESPRIT \cite{VanTrees2002,mskaesprit} and root‑MUSIC to be employed to estimate the DOAs \cite{evans1982,pillai1989,Schmidt1986,Barabell1983-2}.

Coarray processing later extended smoothing concepts to SLAs, revealing that FSS on the vectorized covariance can resolve more sources than physical sensors, generalizing subspace methods such as MUSIC to deal with coarray data \cite{Pal2010-1,wang2017,leite2024,leite2022}. Yet these schemes still adopt a fixed smoothing window size (FWS) to process the coarray data, including the deterministic noise in all spatially-smoothed coarrays (SS-CA). Overcoming that limitation motivates the variable‑window size scheme introduced in this paper.

In this work, we introduce a variable-window size spatial-smoothing (VWS) scheme for coarray DOA estimation with SLAs. By letting the smoothing aperture shrink by an integer parameter, the method replaces perturbed with unperturbed (noise-variance-free or deterministic-noise-free) outer products and thus enhances estimation accuracy at reduced computational cost. We derive two algorithms, namely VWS-CA-MUSIC and its root variant VWS-CA-rMUSIC. The computational complexity scales with the cube of the reduced aperture, offering relevant savings over fixed-window counterparts. Simulations show the performance increase relative to standard coarray MUSIC.

\section{Measurement Model and Problem Formulation}\label{sec:systemModel}
Consider a sensor array with a total of $N$ elements disposed linearly, $T$ data snapshots and $D$ impinging uncorrelated sources coming from far field. In this case, the received signal is given by
\begin{equation}\label{eq:data_model_param}
    \v{x}_{\set{S}}(t)=\v{A}_{\set{S}}(\v{\theta})\v{s}(t)+\v{n}_{\set{S}}(t) \inc{N},~t\in\sqb{T}
\end{equation}
where $t$ is the time index, $\sqb{\v{A}_\set{S}(\v{\theta})}_{k,d}=\exp(-j\pi n_k\theta_d)$ is the $(k,d)$-th steering matrix element, $n_k\in{\set{S}}$ is the $k$-th sensor position, $\set{S}$ means the set of sensor positions as an integer multiple of the minimum intersensor distance, half of the carrier wavelength, $\v{s}(t)\inc{D}$ is the source signal, and $\v{n}_{\set{S}}(t)\inc{N}$ denotes the noise vector. The sources come from directions given by $\v{\theta}\in [-1,1)^{D}$ (sines of the DOAs). The noise and the source signals are drawn from a circularly complex multivariate Gaussian distribution. The noise is spatially and temporally white and uncorrelated from the sources. Any noise and signal components are uncorrelated, regardless of the time and spatial indices. 

The covariance matrix for the array is given by
\begin{meq}\label{eq:cov_mat_sub}
    \v{R}_{\set{S}} & = \expt{\v{x}_{\set{S}}(t)\v{x}_{\set{S}}^H(t)}\\ 
    & = \v{A}(\v{\theta})\v{R}_s\v{A}^H(\v{\theta})+\sigma_n^2\v{I}_{N}.
\end{meq}
By applying the vectorization operator in (\ref{eq:cov_mat_sub}), we obtain
\begin{equation}\label{eq:vec_op}
  \v{z} = \prt{\v{A}_{\set{S}}^{\ast}(\v{\theta})\circ\v{A}_{\set{S}}(\v{\theta})}\v{p}+\sigma_n^2\bar{\v{i}},
\end{equation}
with $\circ$ denoting the Khatri-Rao product, $\v{p}=\opn{diag}(\v{R}_s)\inc{D\times 1}$ is the vector of source powers, $\bar{\v{i}}=\sqb{\v{e}_1^T,\ldots,\v{e}_{N}^T}^T$ is the vectorization of the identity matrix, and $\v{e}_i$ is the $i$-th basis vector from a canonical algebraic convention. For finite-snapshot scenarios, averaging the repeated rows in $\v{A}_{\set{S}}^{\ast}(\v{\theta})\circ\v{A}_{\set{S}}(\v{\theta})$ and sorting the coarray elements in ascending order, we have
\begin{equation}\label{eq:scaa}
  \v{x}_{\set{D}}=\v{A}_{\set{D}}(\v{\theta})\v{p}+\sigma_n^2\v{i},
\end{equation}
where $\set{D}$ is the coarray set, $\sqb{\v{A}_{\set{D}}(\v{\theta})}_{m,d}=\exp(j\pi n_m \theta_d)$ with $n_m\in \set{D}$, $\v{x}_{\set{D}}\inc{|\set{D}|}$ is the coarray received signal and $\v{i}\in\{0,1\}^{|\set{D}|}$ is an all-zero vector with the exception of a $1$ in its half position (element $(|\set{D}|+1)/2$). Our goal is to estimate the source directions $\v{\theta}$ from (\ref{eq:scaa}).

\section{Variable Window Size Spatial Smoothing for Coarray DOA estimation}\label{sec:vws_coarray_doaest}

In this section, we briefly introduce the variable window size scheme for coarray processing, which stands out as a more accurate approach using coarray data, SLAs and uncorrelated sources. In coarray MUSIC \cite{Pal2010-1}, forward spatial smoothing (FSS) is used in the coarray received signal $\v{x}_{\set{D}}$ with spatial smoothing coarrays (SS-CA) of fixed size equal to $G=(\text{UDOF}+1)/2$, resulting in
\begin{equation}\label{eq:ss_coarray}
	\v{R}_{\set{D}}^{\text{SS}} =\frac{1}{G}\sum_{g=1}^{G}\v{x}_{\set{D}}^g\prt{\v{x}_{\set{D}}^g}^H
\end{equation}
where $\text{UDOF}$ is the number of \emph{uniform} DOF and corresponds to the number of distinct contiguous covariance lags in the covariance matrix \cite{Liu2016-2}. Moreover, we also define 
\begin{equation}
    \v{x}_{\set{D}}^g\triangleq \v{J}_g\v{x}_{\set{D}}\inc{G}
\end{equation}
where $\v{x}_{\set{D}}^g$ represents the $g$-th overlapping SS-CA and $\v{J}_g$ is defined as
\begin{equation}
	\v{J}_g = \sqb{\v{0}_{G\times (G-g)}~\v{I}_G~\v{0}_{G\times(g-1)}}
\end{equation}
It is known from the literature that (\ref{eq:ss_coarray}) presents a signal and noise subspaces, which can be found through eigenvalue decomposition (EVD), allowing algorithms like MUSIC to be employed \cite{Pal2010,Pal2010-1}. Since $G=\mathcal{O}(N^2)$ $\prt{\text{UDOF}=\mathcal{O}(N^2)}$, in most scenarios we have $G\gg D$, the number of sources, indicating that the SS-CA aperture could be much larger than the number of detected sources. This intuition motivates a variant of the coarray MUSIC algorithm specifically designed for scenarios in which the number of sources is much smaller than the available degrees of freedom in the coarray domain.

This strategy will be demonstrated to improve the estimation accuracy of subspace-based algorithms such as MUSIC by producing a more refined averaging process for the smoothing average, as the number of rank-one (outer product) matrices to be averaged increases. Mathematically, we can write
\begin{equation}\label{ç}
	M = \text{UDOF}-P+1
\end{equation}
where $P$ is the number of \emph{Spatially-Smoothed Coarrays} (SS-CA) and $M$ is their aperture. From now on, it is assumed that the variable $G$ has a fixed value of $(\text{UDOF}+1)/2$, but the values of $M$ and $P$ can vary, i.e., $P$ can increase and $M$ is shrunk by a given factor. Define the difference between M and G as $G-M=a$, where $a\in \set{N}_{+}$  indicates the shrinkage factor for the smoothing window. In case of coarray MUSIC, $a=0$.

As the parameter $a$ increases, the SS-CA aperture $M$ decreases, while the number $P$ of submatrices of order $M=G-a$ increases. This alters the trade-off between coarray resolution and RMSE performance. When the sources are sufficiently separated and the DOF significantly exceed the number of sources, this leads to a more effective setting than that achieved by coarray MUSIC.

Nevertheless, when $P$ and the aperture $M$ are no longer matched, it is essential to ensure that spatial smoothing in the coarray domain remains viable for signal and noise subspace decomposition. Specifically, the rank of the source covariance matrix—formed via the outer product of the SS-CA must be preserved to maintain the structural integrity of the signal and noise subspace directions as represented by their eigenvectors.

From now on, the equations will be developed considering $a>0$. The case $a=0$ corresponds to plain coarray MUSIC (CA-MUSIC). The spatial smoothing sum is given by
\begin{equation}\label{eq:coarray_ss}
	\widetilde{\v{R}}_{\set{D}}^{\text{SS}} = \frac{1}{P}\sum_{p=1}^P \widetilde{\v{R}}_{\set{D}^p}
\end{equation}
where
\begin{equation}
	\widetilde{\v{R}}_{\set{D}^p} = \widetilde{\v{x}}_{\set{D}^p}\widetilde{\v{x}}_{\set{D}^p}^H
\end{equation}
and
\begin{equation}\label{eq:indss-sca}
	\widetilde{\v{x}}_{\set{D}^p} =
	\begin{cases}
		\v{A}_{\set{D}^r}\v{\Phi}^{p-a-1}\v{p}+\sigma_n^2\v{e}_{p-a}\quad\text{if } \\
		  a+1 \leq p \leq M+a\\
		\v{A}_{\set{D}^r}\v{\Phi}^{p-a-1}\v{p}\quad\text{if } 1\leq p \leq a \text{ or }\\
		M+a+1 \leq p \leq M+2a \\
	\end{cases}
\end{equation}
where $\v{A}_{\set{D}^r}$ is the coarray manifold corresponding to the $(a+1)$-th subarray, taken as reference. By making the change of variable $i=p-a-1$, we can rewrite (\ref{eq:indss-sca}) as
\begin{equation}\label{eq:indss-sca-2}
	\v{y}_i =
	\begin{cases}
		\v{A}_{\set{D}^r}\v{\Phi}^{i}\v{p}+\sigma_n^2\v{e}_{i+1}\quad \text{if } 0 \leq  i \leq M-1,      \\
		\v{A}_{\set{D}^r}\v{\Phi}^{i}\v{p}\quad\text{if } -a\leq i \leq -1\text{ or }\\
		M \leq i \leq M+a-1 \\
	\end{cases}
\end{equation}
with $\v{y}_i=\widetilde{\v{x}}_{\set{D}^{i+a+1}}$. In (\ref{eq:indss-sca-2}), the first case corresponds to $M=G-a$ perturbed (noise-variance-dependent) SS-CA, whereas the second case corresponds to $2a$ \emph{unperturbed} SS-CA. Consequently, this results in a total of $G-a+2a=G+a$ SS-CA, compared to $G$ perturbed SS-CA in the conventional CA-MUSIC algorithm. Then, the VWS scheme increased the number of covariance matrices to be averaged.

The final spatial smoothing matrix can be written as 
\begin{equation}\label{eq:rss_coarray_2}
	\widetilde{\v{R}}_{\set{D}}^{\text{SS}} = \frac{1}{P}\prt{\v{R}_1^2+\v{R}_2^2}	
\end{equation}
where the full-rank matrix $\v{R}_1$ is defined as $\v{R}_1=\v{A}_{\set{D}^r}\opn{diag}(\v{p})\v{A}_{\set{D}^r}^H+\sigma_n^2\v{I}$ and $\v{R}_2^2 = \v{A}_{\set{D}^r}\v{B}\v{B}^H\v{A}_{\set{D}^r}^H$, consisting of an additive low-rank term on $\v{R}_1^2$. The matrix $\v{B}$ is defined as
\begin{meq}
\v{B} & = \opn{diag}\prt{\v{p}}\times\\
	&\begin{bmatrix}
		\omega_1^{-a} & \omega_1^{-a+1} & \cdots & \omega_1^{-1} & \omega_1^{G-a} & \cdots & \omega_1^{G-1} \\
		\omega_2^{-a} & \omega_2^{-a+1} & \cdots & \omega_2^{-1} & \omega_2^{G-a} & \cdots & \omega_2^{G-1} \\
		\vdots        & \vdots          & \cdots & \vdots        & \vdots         & \cdots & \vdots         \\
		\omega_D^{-a} & \omega_D^{-a+1} & \cdots & \omega_D^{-1} & \omega_D^{G-a} & \cdots & \omega_D^{G-1} \\
	\end{bmatrix}
\end{meq}

Moreover, it can be demonstrated that $\v{R}_2^2$ does not disrupt the signal and noise subspaces of $\v{R}_1$, thus preserving the subspace structure. This is crucial because it allows us to use subspace-based algorithms such as MUSIC to find the DOAs.

We also add that the smallest value of the parameter $a$ that ensures $\v{R}_2^2$ achieves full rank is $D$. In contrast, if the parameter $a$ is smaller than $D$, as it is usually the case, then the matrix $\v{R}_2^2$ will have a rank of $a$, indicating that it is rank-deficient by $D-a$. 

We observe that even in the scenario where $\v{R}_2^2$ retains full rank, it continues to be a low-rank additional term of the rank-$M$ matrix $\v{R}_1^2$. Moreover, the parameter $a$ must obey the condition $a<\prt{\text{UDOF}+1-2D}/2$, implying that excessively reducing the window size may compromise the minimum dimensionality required for subspace-based algorithms to distinguish signal from noise subspaces. The VWS-CA-MUSIC algorithm is summarized in Algorithm~\ref{alg:vws_ca_music}.

\begin{algorithm}
	\DontPrintSemicolon
	\SetKwInOut{Inp}{Input}
	\SetKwInOut{Out}{Output}
	\Inp{Array geometry $\mathbb{S}$, data matrix $\widehat{\v{X}}_{\set{S}}$, grid $\v{\theta}^g\in (-1,1]^g$, number of sources $D$, window compression parameter $a$}
	\BlankLine
	$\widehat{\v{R}}_{{\mathbb{S}}}\leftarrow (1/T)\widehat{\v{X}}_{\set{S}}\widehat{\v{X}}_{\set{S}}^H$\;
	$\widehat{\v{z}} \leftarrow \opn{vec}\prt{\widehat{\v{R}}_{{\mathbb{S}}}}$\;
	Average the repeated elements in $\widehat{\v{z}}$ and sort the corresponding coarray locations to obtain $\widehat{\widetilde{\v{x}}}_{\set{D}}$\;
	$\widehat{\widetilde{\v{R}}}_{\set{D}}^{\text{SS}} \leftarrow (1/P)\sum_{p=1}^{P}\widehat{\widetilde{\v{x}}}_{\set{D}^p}\widehat{\widetilde{\v{x}}}_{\set{D}^p}^H$\;
	Diagonalize $\widehat{\widetilde{\v{R}}}_{\set{D}}^{\text{SS}}$ through an EVD to find $\widehat{\widetilde{\v{U}}}_{N}$\;
	\For{$i\leftarrow 1$ \KwTo $g$}{
		$\widehat{\widetilde{P}}(\theta_i^g)=\sqb{\sqb{\widetilde{\v{a}}_{\set{D}}^{\text{SS}} (\theta_i^g)}^H\widehat{\widetilde{\v{U}}}_{N}\widehat{\widetilde{\v{U}}}_{N}^H\widetilde{\v{a}}_{\set{D}}^{\text{SS}} (\theta_i^g)}^{-1}$\;
	}
	Find the peaks in $\widehat{\widetilde{P}}(\theta_i^g)$ and take the corresponding angles as the estimated DOAs\;
	\BlankLine
	\caption{VWS-CA-MUSIC}\label{alg:vws_ca_music}
	\Out{Estimated DOAs $\widehat{\v{\theta}}$}
\end{algorithm}
Based on the subspace properties of the VWS algorithm, a search-free root version of it can also be derived and is presented in Algorithm~\ref{alg:vws_ca_rmusic}.

\begin{algorithm}
	\DontPrintSemicolon
	\SetKwInOut{Inp}{Input}
	\SetKwInOut{Out}{Output}
	\Inp{Array geometry $\mathbb{S}$, data matrix $\widehat{\v{X}}_{\set{S}}$, grid $\v{\theta}^g\in (-1,1]^g$, number of sources $D$, window compression parameter $a$}
	\BlankLine
	$\widehat{\v{R}}_{{\mathbb{S}}}\leftarrow (1/T)\widehat{\v{X}}_{\set{S}}\widehat{\v{X}}_{\set{S}}^H$\;
	$\widehat{\v{z}} \leftarrow \opn{vec}\prt{\widehat{\v{R}}_{{\mathbb{S}}}}$\;
	Average the repeated elements in $\widehat{\v{z}}$ and sort the corresponding coarray locations to obtain $\widehat{\widetilde{\v{x}}}_{\set{D}}$\;
	$\widehat{\widetilde{\v{R}}}_{\set{D}}^{\text{SS}} \leftarrow (1/P))\sum_{p=1}^{P}\widehat{\widetilde{\v{x}}}_{\set{D}^p}\widehat{\widetilde{\v{x}}}_{\set{D}^p}^H$\;
	Diagonalize $\widehat{\widetilde{\v{R}}}_{\set{D}}^{\text{SS}}$ through an EVD to find $\widehat{\widetilde{\v{U}}}_{N}$\;
	Find the roots of $Q_{\set{D}}(z) = \v{f}_{\set{D}}^T(1/z)\widehat{\widetilde{\v{U}}}_{N}\widehat{\widetilde{\v{U}}}_{N}^H\v{f}_{\set{D}}(z)$\;
	Select the $D$ roots inside and closest to the unit circle as $\widehat{z}_d$\;
	Estimate the DOAs according to $\widehat{\theta}_d = \prt{\angle \widehat{z}_d}/\pi$\;
	\BlankLine
	\caption{VWS-CA-rMUSIC} \label{alg:vws_ca_rmusic}
	\Out{Estimated DOAs $\widehat{\v{\theta}}$}
\end{algorithm}

In coarray MUSIC, the dominant computational cost arises from the eigenvalue decomposition (EVD) of the spatially smoothed covariance matrix in (\ref{eq:coarray_ss}), whose complexity scales as \(\mathcal{O}(M^3)\).\ In the proposed VWA schemes, the effective matrix dimension is reduced to \(M^\prime = M - a\), since the spatially smoothed coarrays (SS-CAs) are formed over a shortened aperture. Because the EVD cost depends cubically on the matrix size, the VWA-based variants lead to a reduced overall complexity. This reduction becomes more pronounced for array geometries that offer a larger number of degrees of freedom, as they provide sufficient flexibility to increase \(a\) and apply stronger shrinkage without distorting the underlying signal and noise subspace structures.

\section{Simulation Results}\label{sec:results}

In this section, we use numerical simulations to evaluate the performance of the VWS schemes for the proposed algorithms, namely VWS-CA-MUSIC and VWS-CA-rMUSIC. This demonstrates the gains achieved by compressing the smoothing window due to the exploitation of the additive low-rank term in the smoothed matrix. We have used $T=1,000$ snapshots, $D=3$ uncorrelated sources located at $\v{\theta}=\sqb{-0.8,0,0.8}$ and $N=8$ physical sensors, except for the last experiment, for which we have used $D=5$ sources. The results were averaged over $10,000$ trials. The SLA geometries consider Two-Level Nested Arrays (NAQ2) \cite{Pal2010-1}, 2nd-Order Super Nested Arrays (SNAQ2) \cite{Liu2016}, and Minimum Redundancy Arrays (MRA) \cite{Moffet1968}. 

\begin{figure}
	\centering
	\includegraphics[width=.4\textwidth]{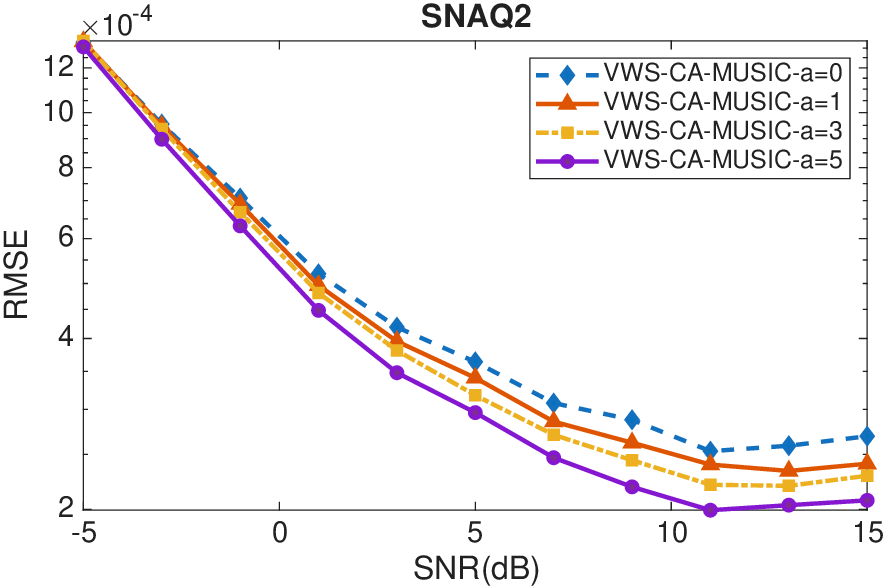}
	\caption{VWS-CA-MUSIC: RMSE comparison for multiple values of $a$ against SNR. SNAQ2 geometry. $a=0$ corresponds to the CA-MUSIC (coarray MUSIC) algorithm with a fixed window size \cite{Pal2010-1}.}\label{fig:vws_ca_music_snr}
\end{figure}

\begin{figure}
	\centering
	\includegraphics[width=.4\textwidth]{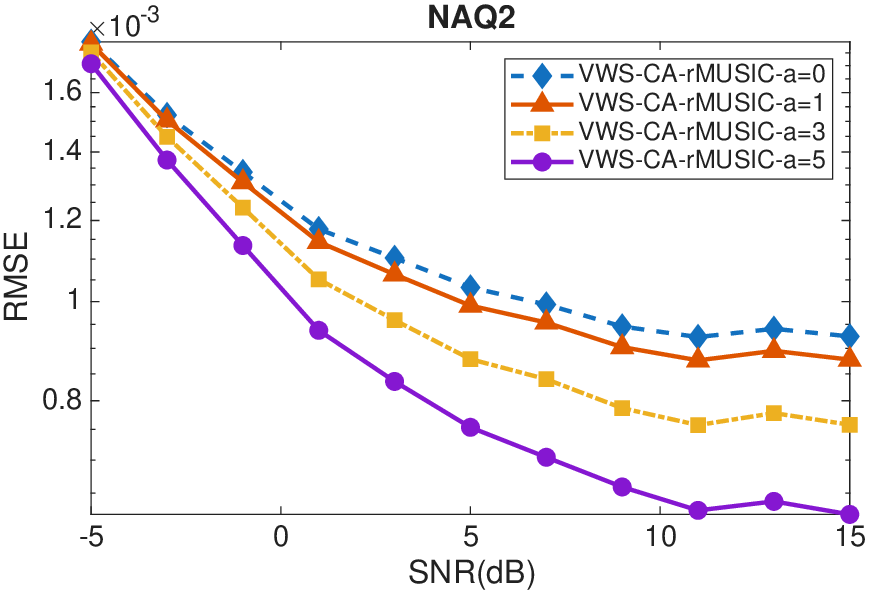}
	\caption{VWS-CA-rMUSIC: RMSE comparison for multiple values of $a$ against SNR. NAQ2 geometry.}\label{fig:vws_ca_rmusic_snr}
\end{figure}

From Fig.~\ref{fig:vws_ca_music_snr} and Fig.~\ref{fig:vws_ca_rmusic_snr}, it is clear the performance gain achieved by VWS schemes for both geometries: SNAQ2 and NAQ2, with the gains for NAQ2 and VWS-CA-rMUSIC being more prominent. Indeed, the aperture shrinkage increases the RMSE performance of the processing using the coarray data.

\begin{figure}
	\centering
	\includegraphics[width=.4\textwidth]{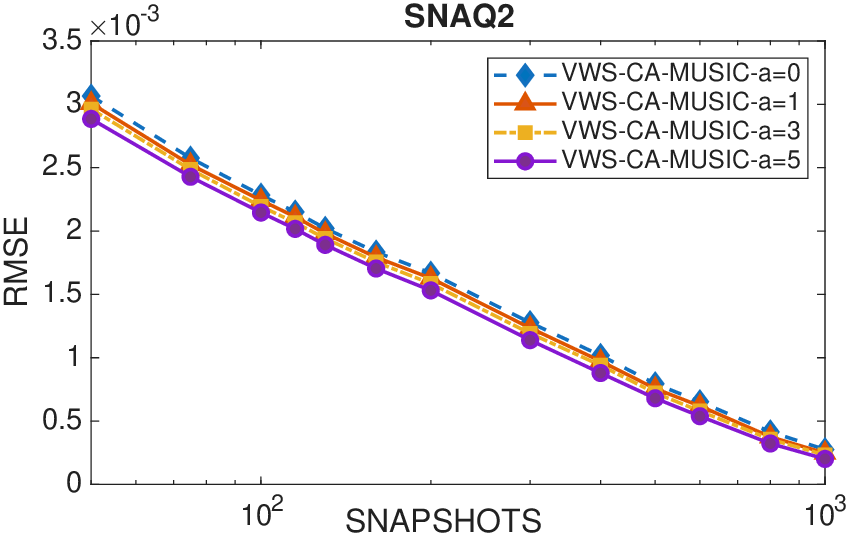}
	\caption{VWS-CA-MUSIC: RMSE against snapshots. SNAQ2 geometry.}\label{fig:vws_ca_music_snaps}
\end{figure}

\begin{figure}
	\centering
	\includegraphics[width=.4\textwidth]{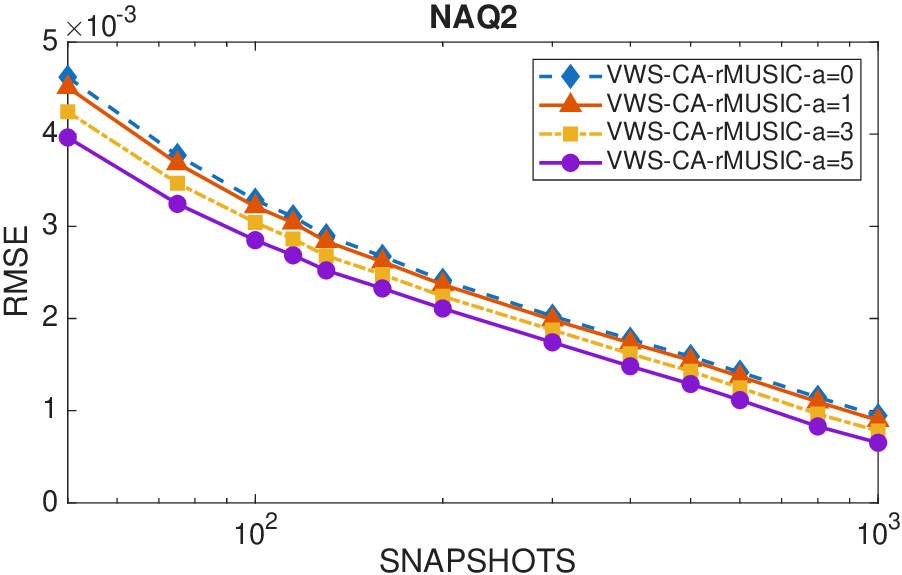}
	\caption{VWS-CA-rMUSIC: RMSE against snapshots. NAQ2 geometry.}\label{fig:vws_ca_rmusic_snaps}
\end{figure}

In addition to the previous scenarios, we also analyze the algorithm’s performance as the amount of available data increases successively. Fig.~\ref{fig:vws_ca_music_snaps} and Fig.~\ref{fig:vws_ca_rmusic_snaps} depict this behavior for an \(\text{SNR}=10~\text{dB}\), using the same parameter settings adopted in the previous case. This analysis is intended to highlight the algorithm’s performance when fewer data samples are available.

In this scenario, the performance gains provided by NAQ2 in combination with VWS-CA-rMUSIC are substantially larger than those obtained with SNAQ2 and VWS-CA-MUSIC under identical parameter settings. The results further indicate that the variable window size strategy significantly improves DOA estimation accuracy in data-limited regimes, thereby confirming the robustness of the proposed method. In addition, the RMSE gains (difference between the curves) do not appear to increase significantly with additional snapshots for different values of $a$, suggesting that the relative RMSE between the algorithms is dominated by the spatial-smoothing window aperture, which is controlled by $a$, rather than by the amount of available data.

\begin{figure}
	\centering
	\includegraphics[width=.4\textwidth]{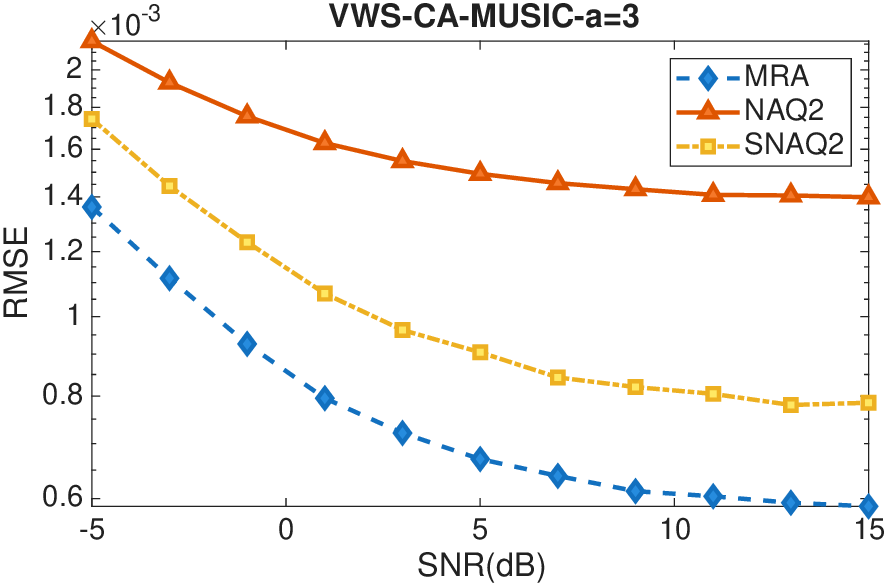}
	\caption{RMSE against SNR comparing multiple geometries: NAQ2, SNAQ2, and MRA using VWS-CA-MUSIC algorithm.}\label{fig:vws_ca_music_snr_geom}
\end{figure}

Additionally, a numerical experiment is conducted to compare the RMSE performance of VWS-CA-MUSIC as a function of SNR, for a shrinkage parameter $a = 3$ and multiple array geometries. In this specific case, we increased the number of sources to $D=5$ located at $\v{\theta} = \sqb{-0.8,-0.4,0,0.4,0.8}$. Fig.~\ref{fig:vws_ca_music_snr_geom} displays the resulting RMSE curves. It can be observed that the MRA configuration attains the lowest RMSE over the entire SNR range. This behavior is consistent with the fact that MRA provides the largest virtual aperture and therefore incurs the smallest effective aperture reduction when the spatial-smoothing window is shrunk by a factor of $a$. The performance gap among the three geometries becomes more pronounced as the SNR increases.

\section{Conclusions}\label{sec:conclusions}

In this paper, the VWS-CA-MUSIC and VWS-CA-rMUSIC (search-free) algorithms are proposed, which exploit spatial smoothing techniques to enhance coarray-based DOA estimation for sparse linear array configurations. The proposed framework incorporates an adjustable spatial-smoothing window, yielding substantial gains in estimation accuracy for sparse arrays. In addition, an upper bound on the shrinkage factor is derived to ensure that the structures of the signal and noise subspaces are preserved under the spatial-smoothing-based shrinkage.

\bibliographystyle{IEEEtran}
\bibliography{mybib}

\end{document}